\documentclass[letterpaper, 10 pt, conference]{ieeeconf} 
\IEEEoverridecommandlockouts                              
\usepackage{times}

\usepackage{subfiles}
\usepackage{multicol}
\usepackage[bookmarks=true]{hyperref}
\usepackage{balance}
\usepackage{hyperref}
\usepackage{graphicx}
\usepackage{amssymb}
\usepackage{amsmath}
\usepackage{amsfonts}
\usepackage{caption}
\usepackage{algorithm}
\usepackage{algorithmicx}
\usepackage[noend]{algpseudocode}
\usepackage{graphicx,caption}
\usepackage{amsmath,amssymb,stmaryrd,mathtools}
\usepackage{url}
\usepackage{subcaption}
\usepackage{booktabs}
\usepackage{multirow}
\usepackage{float}
\usepackage{amsmath}
\usepackage{adjustbox}
\DeclareMathOperator*{\argmax}{arg\,max}
\DeclareMathOperator*{\argmin}{arg\,min}
\hypersetup{
    colorlinks=true,
    linkcolor=blue,
    citecolor=blue,
    filecolor=magenta,      
    urlcolor=cyan,
    hyperindex=True,
 urlcolor=blue}
\usepackage{wrapfig}

\newcommand{\HB}[1]{}

\newtheorem{theorem}{Theorem}[section]

\newtheorem{lemma}[theorem]{Lemma}

\title{\bf LEAF: Latent Exploration Along the Frontier}

  \author{ Homanga Bharadhwaj$^{1}$, Animesh Garg$^{1}$, and Florian Shkurti$^{1}$
\thanks{$^{1}$ Authors are with Department of Computer Science, University of Toronto, Canada
        {\tt\small homanga@cs.toronto.edu}}%
} 
    
\begin{document}
    
    \maketitle

\begin{abstract}
Self-supervised goal proposal and reaching is a key component for exploration and efficient policy learning algorithms. Such a self-supervised approach without access to any oracle goal sampling distribution requires deep
exploration and commitment so that long horizon plans can be efficiently discovered. In this paper, we propose an
exploration framework, which learns a dynamics-aware manifold of reachable states. For a goal, our proposed
method deterministically visits a state at the current frontier of reachable states (commitment/reaching) and then
stochastically explores to reach the goal (exploration). This allocates exploration budget near the frontier of the
reachable region instead of its interior. We target the challenging problem of policy learning from initial and
goal states specified as images, and do not assume any access to the underlying ground-truth states of the robot
and the environment. To keep track of reachable latent states, we propose a distance-conditioned reachability
network that is trained to infer whether one state is
reachable from another within the specified latent space
distance. Given an initial state, we obtain a frontier of
reachable states from that state. By incorporating a
curriculum for sampling easier goals (closer to the start
state) before more difficult goals, we demonstrate that the
proposed self-supervised exploration algorithm, superior
performance compared to existing baselines on a set of
challenging robotic environments.\\ \href{https://sites.google.com/view/leaf-exploration}{https://sites.google.com/view/leaf-exploration}
\end{abstract}
\vspace*{-0.15cm}
\section{Introduction}
\label{sec:intro}  
\vspace*{-0.15cm}
Efficient exploration is one of the central open challenges in Reinforcement Learning (RL), and a key requirement for learning optimal policies, particularly in sparse-reward and long-horizon settings, where the optimization problem is difficult to solve, but easy to specify. We show that discovering policies for  these settings benefits significantly from autonomous goal-setting and exploration, which requires keeping track of what regions of the state-space have been sufficiently explored and can be reached via planning, and what regions remain unknown. In this paper we present an exploration method that keeps track of reachable regions of state space and naturally handles the exploration-exploitation dilemma in sequential decision-making for unknown environments~\cite{varibad,goexplore}, while being practical in real robotics settings. In addition to experimental evaluation, we also show proofs for simplified exploration settings based on expected reaching time analysis. \HB{General introduction}


For learning to plan in \textit{unknown} environments, appropriately trading off exploration i.e. learning about the environment, and exploitation i.e. executing promising actions is the key to successful policy learning. The necessity for this trade-off is more pertinent when we want to train agents to learn autonomously in a self-supervised manner such that they are able to solve goals sampled from an unknown distribution during evaluation. In this setting, during training, the agent must practice reaching self-specified random goals~\cite{rig}. However, trying to specify goals through exact states of objects in the environment will require a combinatorially large representation, for explicitly randomizing different variables of the states~\cite{avi,rig}. On the contrary, raw sensory signals, such as images can be used to specify goals, as has been done in recent goal-conditioned policy learning algorithms~\cite{rig,upn,skew}\HB{In the previous two lines I am trying to motivate the need for learning from images even in simulation. In the real environment it is clear why we want to use images as input - we cannot describe the environment state exactly.}. This paper tackles the question of how should an autonomous agent balance exploration and exploitation for goal-directed robotic tasks while learning directly from images, with no access to the ground-truth parameters of the environment during training. 

\begin{figure}
    \centering
   \includegraphics[width=\columnwidth]{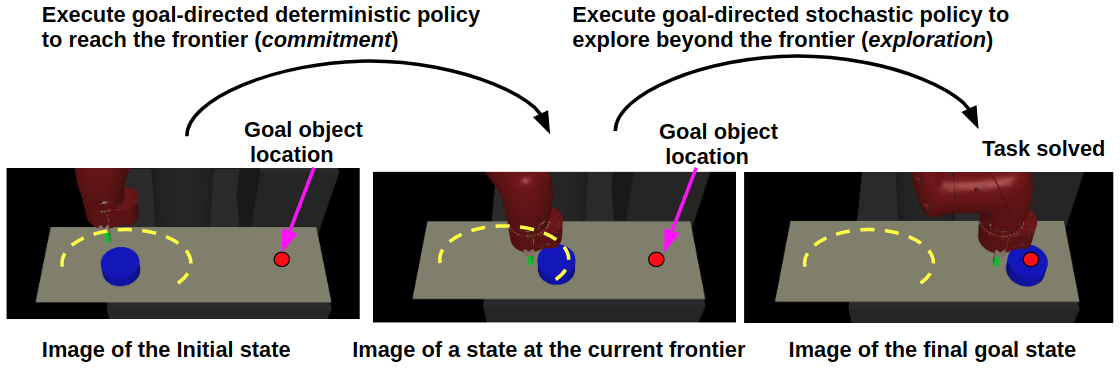}
    \caption{LEAF prioritizes reaching the current frontier of reachable states with a deterministic policy, and performing goal-directed exploration beyond it with a stochastic policy. Here, the yellow region corresponds to puck positions for the currently reachable frontier corresponding to the shown initial state.} \label{fig:firstfig}
    \vspace*{-0.8cm}
\end{figure}

The main idea of our approach is to maintain and gradually expand the frontier of reachable states by trying to reach the current frontier directly without exploring and exploring from beyond the current frontier in order to reach the goal (Fig.~\ref{fig:firstfig}). In order to achieve this, we encourage the agent to sample goals closer to the start state(s) before sampling goals that are farther away, by appropriately querying the reachability network. 
Since planning directly in the pixel-space is not meaningful~\cite{upn,rig}, we learn a latent representation from images, and plan in the latent space. We train a reachability network that learns whether a latent state is currently reachable from another latent state in $k$ time-steps ($k$ varies from $1$ to the maximum length of the horizon), and use this network to compute the current frontier of reachable states given a random initial start state during training. While training the goal-conditioned policy to reach a goal latent state (corresponding to a goal image), we first execute the deterministic policy trained so far to reach a latent state at the frontier (\textit{commitment}), and then execute the stochastic policy from beyond the frontier to reach the goal (\textit{exploration}). We refer to this procedure of commitment to reach the frontier and exploring beyond it as \textit{committed exploration} required for solving temporally abstracted tasks. \HB{The proposed approach. Every episode in training has two phases - execute deterministic policy to reach frontier. Explore beyond frontier while trying to reach goal. I don't mean to say the second phase here is "deep exploration". The overall scheme of having this two stage approach is enforcing deep exploration - to discover temporally abstracted winning sequences of actions that solve the task - this process happens completely at the end of training, when the reachable frontier = real frontier.}


This paper makes the following contributions:
\begin{enumerate}
 \item We propose a dynamics-aware latent space reachability network.
 \item We devise a policy learning scheme using the reachability network that combines deterministic and stochastic policies for committed exploration. 
 \item We empirically evaluate the proposed approach  in a range of both image-based, and state-based simulation environments, and in a real robot environment. We demonstrate an improvement of $20\%$ on average compared to state-of-the-art exploration baselines. 
\end{enumerate}

\vspace*{-0.15cm}
\section{Related Works}
\vspace*{-0.15cm}

\noindent \textbf{Vision-based RL for Robotics.} 
Recent works have investigated vision-based learning of RL policies for specific robotic tasks like grasping~\cite{grasping1,grasping2}, pushing~\cite{pushing2,pushing1}, and navigation~\cite{navigation2,navigation1}, however relatively few prior works have considered the problem of learning policies in a completely self-supervised manner~\cite{rig,skew}. A recent paper~\cite{avi} considers RL directly from images, for general tasks without engineering a reward function, but involves a human in the loop. Other prior works learn latent representations for RL in an unsupervised manner which can be used as input to the policy, but they either require expert demonstration data~\cite{upn,sim2realmanga}, or assume access to ground-truth states and reward functions during training~\cite{worldmodels,ground2,ground3}. RIG~\cite{rig} proposes a general approach of first learning a latent abstraction from images, and then planning in the latent space, but requires data collected from a randomly initialized policy to design parametric goal distributions.  In addition, either approaches do not guarantee any notion of exploration or coverage that is required for solving difficult temporally-extended tasks from minimal supervision - in particular with goals specified as raw images.

\noindent \textbf{Goal-conditioned Policy Learning.}
While traditional model-free RL algorithms are trained to succeed in single tasks~\cite{td3,sac,ppo,trpo}, goal-conditioned policy learning holds the promise of learning general-purpose policies that can be used for different tasks specified as different goals~\cite{kaelbling1993learning,her,rig,arc}. However, most goal-conditioned policy learning algorithms make the stronger assumption of having access to an oracle goal sampling distribution during training~\cite{kaelbling1993learning,her,universal,leap}. A recent paper~\cite{arc} proposes an algorithm to create dynamics-aware embeddings of states, for goal-conditioned RL, but it is not scalable to images as goals (instead of ground-truth states).  


\noindent \textbf{Exploration for RL.} Skew-fit~\cite{skew} proposes a novel entropy maximizing objective that ensures gradual coverage in the infinite time limit. In addition, it does not incorporate any notion of commitment during exploration, which is needed to discover complex skills for solving temporally extended tasks in finite time. Go-Explore~\cite{goexplore} addresses this issue to some extent by keeping track of previously visited states in the replay buffer, and at the start of each episode, re-setting the environment to a previously visited state chosen on the basis of domain-specific heuristics, and continuing exploration from there. However, this approach is fundamentally limited for robotics because it assumes the environment is re-settable, which is not always feasible especially for policy learning from images. In addition, the domain-specific heuristics for Montezuma's revenge in the paper~\cite{goexplore} do not readily apply to latent spaces for robotic tasks.

On the contrary, novelty/surprise based criteria have been used in many RL applications like games, and mazes~\cite{curiosity,count,rupam}. 
VIME \cite{vime} considers information gathering behaviors to reduce epistemic uncertainty on the dynamics model. 
A recent paper, Plan2Explore~\cite{plan2explore} incorporates an information gain based intrinsic reward to first learn a world model similar to Dreamer~\cite{dreamer}, without rewards and then learn a policy given some reward function. The main drawbacks of this approach are that it tries to learn a global world model which is often infeasible in complex environments, and requires reward signals from the environment for adaptation. In contrast, our approach can learn to explore using latent rewards alone.


\noindent \textbf{Reachability and Curriculum Learning.} Our method is related to curriculum learning approaches in reinforcement learning \cite{approximate_hjb, curriculum_held, ELMAN199371, openai2019solving, curriculum_bengio, racaniere2019automated} whereby learning easier tasks is prioritized over hard tasks, as long as measures are taken to avoid forgetting behaviors on old tasks. We compute reachability in the latent space and as such our work is distinct from traditional HJ-reachability~\cite{hjb} computations that involve computation of reachable sets of physical robot states by solving Partial Differential Equations (PDEs). 

\noindent \textbf{Frontier-based Exploration for Robot Mapping.} Finally, our method intuitively relates to a large body of classic work on robot mapping, particularly frontier-based exploration approaches \cite{frontier_based_exploration, frontier_based_exploration_multirobot, frontier_based_exploration_holz}.
\section{Problem Setup and Background}
\label{sec:prelim}

We denote the set of observations (images) as $\mathcal{S}$ and the set of goals (images) as $\mathcal{G}$. In this paper we assume the underlying sets for $\mathcal{S}$ and $\mathcal{G}$ to be the same (any valid image can be a goal image~\cite{rig,skew}) but maintain separate notations for clarity. $f_\psi(\cdot)$ denotes the encoder of the $\beta-VAE$~\cite{betavae} that encodes observations $s\sim S$ to latent states $z\sim f_\psi(s)$. The goal conditioned policy given the current latent state $z_t$, and the goal $z_g$ is denoted by $\pi_\theta(\cdot | z_t,z_g)$, and $a_t\sim \pi_\theta(\cdot | z_t,z_g)$ denotes the action sampled from the policy at time $t$. The policy $\pi_\theta(\cdot | z_t,z_g)$ consists of a deterministic component $\mu_\theta(\cdot | z_t,z_g)$, and a noise term $\epsilon$, such that $\pi_\theta(\cdot | z_t,z_g) = \mu_\theta(\cdot | z_t,z_g) + \sigma_\theta\epsilon$, where $\epsilon$ is a Gaussian $\mathcal{N}(0,\mathbf{I})$ in this paper. $ReachNet(z_i,z_j;k)$ denotes the reachability network that takes as input two latent states  $z_i$, and $z_j$ and is conditioned on a reachability integer $k\in[1,..,H]$, where $H$ is the maximum time horizon of the episode. The output of $ReachNet(z_i,z_j;k)$ is either a $0$ denoting $z_j$ is currently not reachable from $z_i$ or a $1$ denoting $z_j$ is reachable from $z_i$. 

We consider the problem setting where a robot is initialized to a particular configuration, and is tasked to reach a goal, specified as an image. The test time goal distribution, from which goal images will be sampled for evaluation is not available during training, and hence the training approach is completely self-supervised. The underlying distribution of observations $\mathcal{S}$ is same as the underlying distribution of goals $\mathcal{G}$ during training, and both are distributions over images from a camera.

\vspace*{-0.2cm}
\section{Our Approach}
\label{sec:approach}
In this section, we discuss the specifics of LEAF .
\begin{figure*}[t]
\centering
   \hspace*{-0.9cm}
        \begin{subfigure}[b]{0.7\textwidth}
               \centering
    \includegraphics[width=0.9\textwidth]{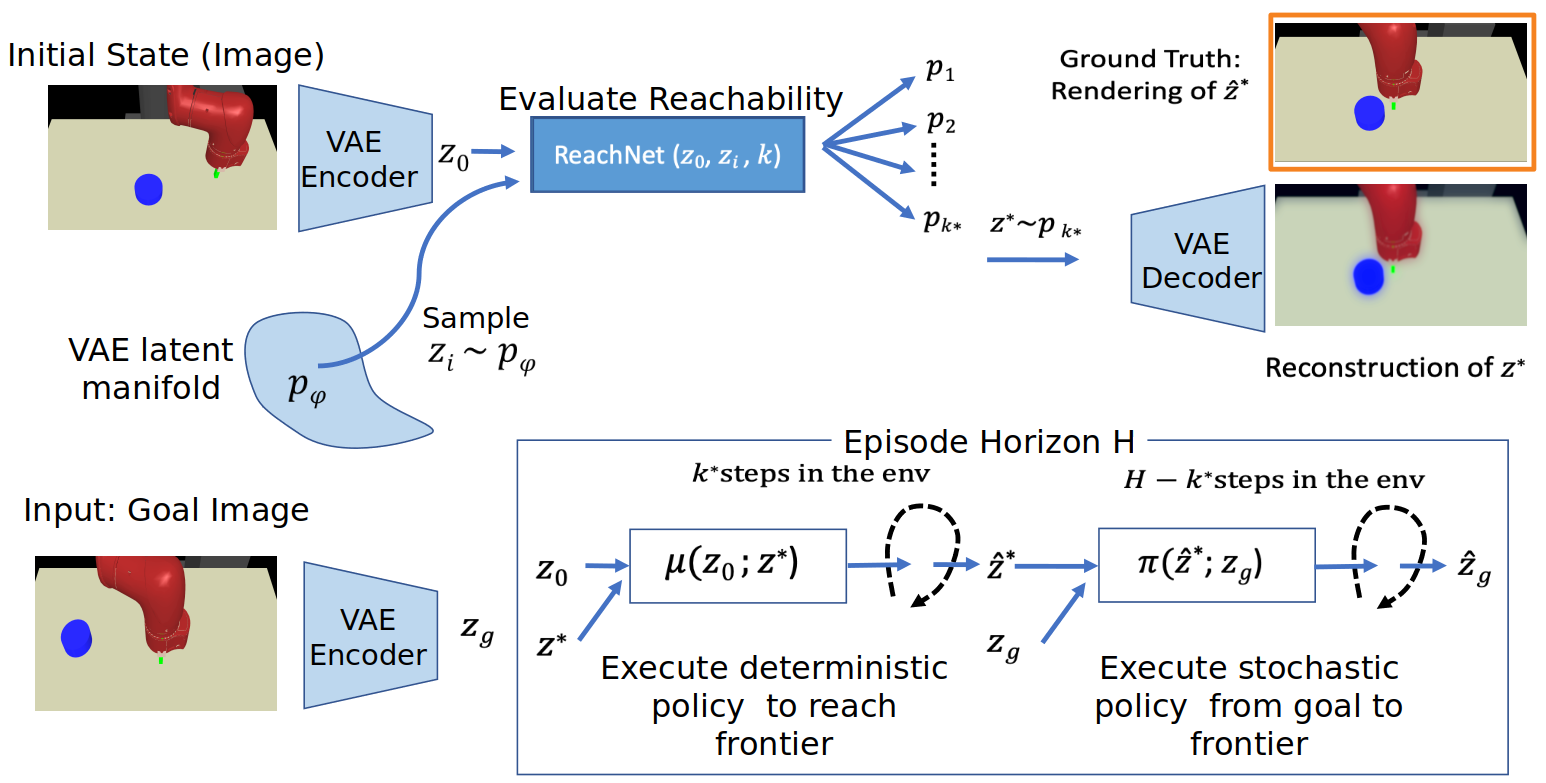}
    \caption{Schematic of the proposed approach}
    \label{fig:reachidea}
        \end{subfigure}
        \hspace*{-0.4cm}
         \begin{subfigure}[b]{0.37\textwidth}
               \centering
    \includegraphics[width=\textwidth]{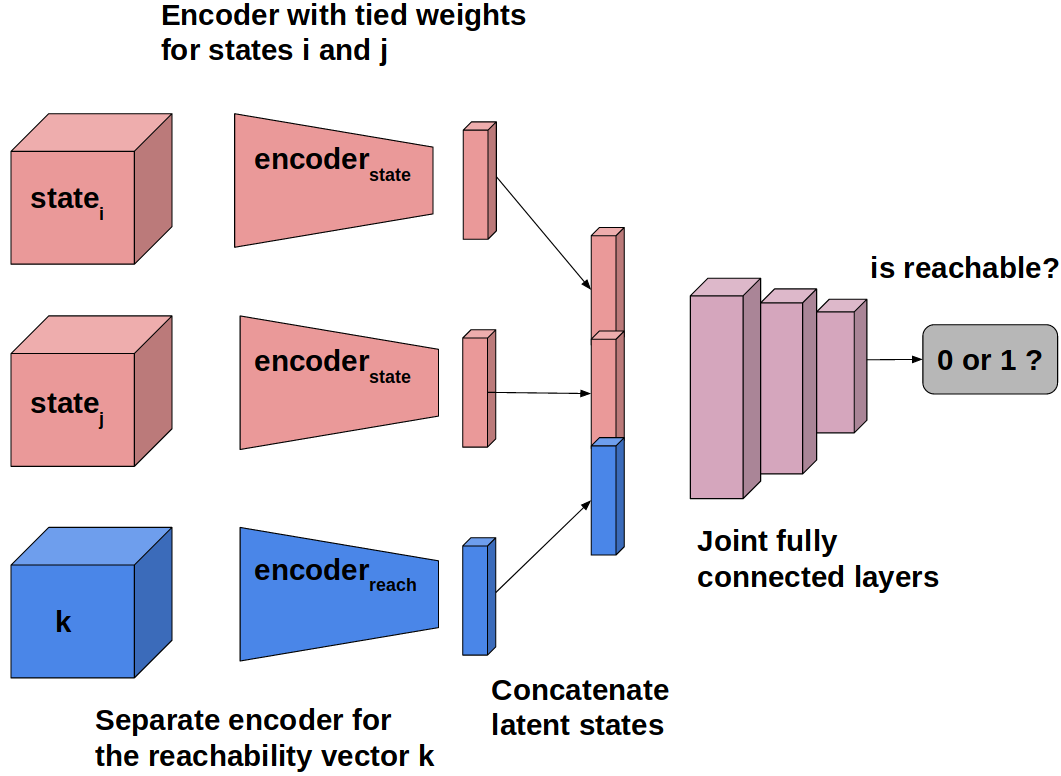}
    \caption{Reachability Network}
    \label{fig:reachnet}
        \end{subfigure}                 
     \caption{(a) An overview of LEAF on the Sawyer Push and Reach task. The initial and goal images are encoded by the encoder of the VAE into latents $z_0$ and $z_g$ respectively. Random states are sampled from the currently learned manifold of the VAE latent states, and are used to infer the current frontier $p_{k^*}$ as described in Section~\ref{sec:reachidea}. The currently learned deterministic policy is used to reach a state in the frontier $z^*$ from the initial latent state $z_0$. After that, the currently learned stochastic policy is used to reach the latent goal state $z_g$. A reconstruction of what the latent state in the frontier $z^*$ decodes to is shown. For clarity, a rendered view of the actual state $\hat{z}^*$ reached by the agent is shown alongside the reconstruction. (b) A schematic of the reachability network used to determine whether the latent state $z_j$ ($state_j$) is reachable from the latent state $z_i$ ($state_i$) in $k$ time-steps. The reachability vector $\mathbf{k}$ consists of a vector of the same size as $z_i$ and $z_j$ with all its elements being $k$. The output $0$ indicates that state $z_i$ is not reachable from state $z_j$.} 
 \label{fig:mainfig}
 \vspace*{-0.4cm}
\end{figure*}

\subsection{The Key Insight of LEAF}
\label{sec:reachidea}
Consider the goal conditioned policy $\pi(\cdot | z_t,z_g)$ with an initial latent state $z_0$, and a goal latent state $z_g$. The key idea of our method is to do committed exploration, by directly going to the frontier of reachable states for $z_0$ by executing the deterministic policy $\mu_\theta(\cdot | z_t,z_{k^*})$ and then executing the exploration policy $\pi_\theta(\cdot | z_t,z_g)$ from the frontier to perform goal directed exploration for the actual goal $z_g$. The frontier defined by $p_{k^*}$ is the empirical distribution of states that can be reliably reached from the initial state, and are the farthest away from the initial state, in terms of the minimum number of timesteps needed to visit them. The timestep index $k^*$ that defines how far away the current frontier is, can be computed as follows:
\begin{equation}
k^* = \argmax_k(IsReachable(z_0;k))
\end{equation}
The binary predicate  $IsReachable(z_0;k)$ keeps track of whether the fraction of states in the empirical distribution of $k-$reachable states $p_k$ is above a threshold $1-\delta$. The value of $IsReachable(z_0;k)$ $\forall k\in[1,..,H]$ before the start of the episode from $z_0$ is computed as:
\begin{equation}
       =\begin{cases}
      True, & \text{if}\ \mathbb{E}_{z\sim p_\phi(z)}ReachNet(z_0,z;k)\geq 1 - \delta \\
      False, & \text{otherwise}
    \end{cases}
  \end{equation}
Here  $p_\phi(z)$ is the current probability distribution of latent states as learned by the $\beta-VAE$~\cite{betavae} with encoder $f_\psi(\cdot)$. $\delta$ is set to 0.2. For computing $\mathbb{E}_{z\sim p_\phi(z)}ReachNet(z_0,z;k)$ above, we randomly sample states $z\sim p_\phi(z)$ from the latent manifold of the VAE (intuition illustrated in the Appendix). After calculating all the predicates $IsReachable(z_0;k)$ $\forall k\in[1,...,H]$ and determining the value of $k^*$, we sample a state $z_{k^*}$ from the empirical distribution of $k^*-$reachable states, $p_{k^*}$, ensuring that it does not belong to any $p_{k}$ $ \forall k< k^*$. Here, the empirical distribution $p_k$ denotes the set of states $z$ in the computation of $\mathbb{E}_{z\sim p_\phi(z)}ReachNet(z_0,z;k)$ for which $ReachNet(z_0,z;k)$ returns a $1$, and  $ReachNet(z_0,z;k')$ returns a $0$, $\forall k'<k$.

In order to ensure that the sampled state $z_{k^*}$ is the closest possible state to the goal $z_g$ among all states in the frontier $p_{k^*}$, we perform the following additional optimization
\begin{equation}
    z_{k^*} = \argmin_{z\in p_{k^*}}||z-z_g||_2
\end{equation}
Here, $||\cdot||_2$ denotes the $L_2$ norm. The above optimization encourages choosing the state in the frontier that is closest in terms of Euclidean distance to the latent goal state $z_g$.

Our method encourages \textit{committed} exploration because the set of states until the frontier $p_{k^*}$ have already been sufficiently explored, and hence the major thrust of exploration should be beyond that, so that new states are rapidly discovered. In our specific implementation, we use SAC~\cite{sac} as the base off-policy model-free RL algorithm for minimizing the Bellman Error in the overall algorithm that is described in the Appendix.

\subsection{Training the Reachability Network}
\label{sec:trainreachnet}
\vspace*{-0.1cm}
The reachability network $ReachNet(z_i,z_j;k)$ is a feedforward neural network with fully connected layers and ReLU non-linearities, the architecture of which is shown in Fig.~\ref{fig:mainfig}, with details in the Appendix. The architecture has three basic components, two encoders ($encoder_{\mathbf{state}}$, $encoder_{\mathbf{reach}}$), one decoder, and one concatenation layer. The latent states $z_i,z_j$ are encoded by the same encoder $encoder_{\mathbf{state}}$ (i.e. two encoders with tied weights, as in a Siamese Network~\cite{siamese}) and the reachability value $k$ is encoded by another encoder $encoder_{\mathbf{reach}}$. In order to ensure effective conditioning of the network $ReachNet$ on the variable $k$, we input a vector of the same dimension as $z_i$ and $z_j$, with all of its values being $k$, $\mathbf{k}=[k,k,...,k]$.

It is important to note that such neural networks conditioned on certain vectors have been previously explored in literature, e.g. for dynamics conditioned policies~\cite{mangapfn}. The three encoder outputs, corresponding to $z_i, z_j,$ and $k$ are concatenated and fed into a series of shared fully connected layers, which we denote as the decoder. The output of  $ReachNet(z_i,z_j;k)$ is a $1$ or a $0$ corresponding to whether $z_j$ is reachable from $z_i$ in $k$ steps or not, respectively.

To obtain training data for $ReachNet$, while executing rollouts in each episode, starting from the start state $z_0$, we keep track of  the number of time-steps needed to visit every latent state $z_i$ during the episode. We store tuples of the form $(z_i,z_j,k_{ij})$ $\forall i<j$ in memory, where $k_{ij} = j-i$, corresponding to the entire episode. Now, to obtain labels $\{0,1\}$ corresponding to the input tuple $(z_i,z_j,k)$, we use the following heuristic:
\begin{equation}
       label(z_i,z_j,k) =\begin{cases}
      1, & \text{if}\ |k_{ij} > \alpha k|\\
      0, & \text{if}\  |k_{ij} < k|
    \end{cases}
  \end{equation}
Here $\alpha$ is a threshold to ensure sufficient separation between positive and negative examples. We set $\alpha=1.3$ in the experiments, and do not tune its value.

\subsection{Implicit Curriculum for Increasing the Frontier}
\label{sec:curricula}
The idea of growing the set of reachable states during training is vital to LEAF. Hence, we leverage curriculum learning to ensure that in-distribution nearby goals are sampled before goals further away.
This naturally relates to the idea of a curriculum~\cite{curriculum}, and can be ensured by gradually increasing the maximum horizon $H$ of the episodes. So, we start with a horizon length of $H=H_0$ and gradually increase it to $H=NH_0$ during training, where $N$ is the total number of episodes.

In addition to this, while sampling goals we also ensure that the chosen goal before the start of every episode does not lie in any of the $p_k$ distributions, where $k\leq k^*$ corresponding to the calculated $k^*$. This implies that the chosen goal is beyond the current frontier, and hence would require exploration beyond what has been already visited.

\vspace*{-0.2cm}
\section{Analysis: Intuition of deep exploration through average case analysis}
\label{sec:analysis}
\begin{figure}[t]
\begin{minipage}[c]{0.49\textwidth}
               \centering
   \includegraphics[width=0.9\textwidth]{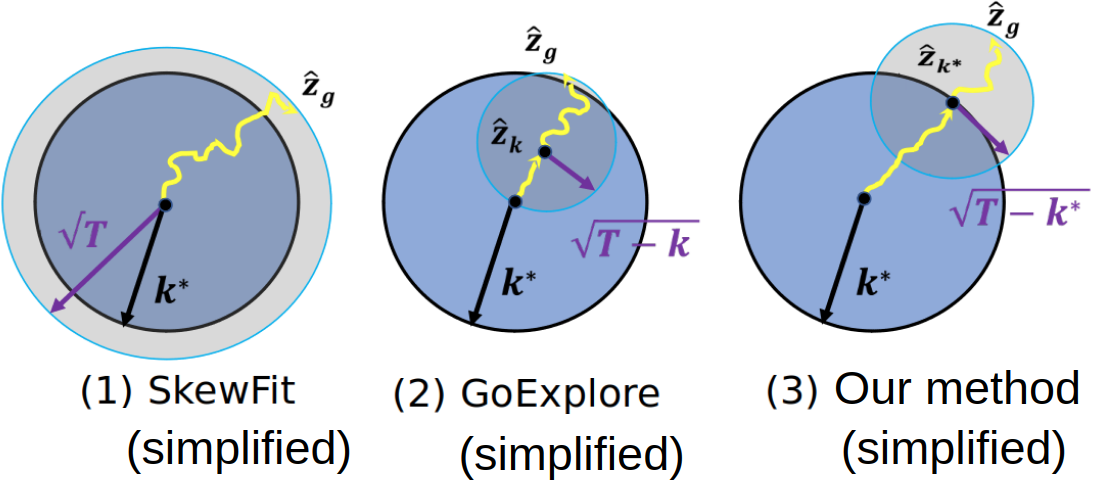}
  \end{minipage} 
  \begin{minipage}[c]{0.45\textwidth}
    \caption{The blue area of radius $k^*$ denotes the already sufficiently explored region such that $k^*$ is the current frontier. $T$ is the max. number of timesteps of the episode.  The simplified models and results are described in Section~\ref{sec:analysis}}
    \label{fig:simpleanalysis}
\end{minipage}    
\vspace*{-0.5cm}
          \end{figure}

\vspace*{-0.45cm}
In this section, we provide formal analysis of our method in a simplified, idealized, setting. Consider a 2D world with a single start location $\mathbf{z}_0$ and and let the goal be denoted by $\mathbf{z}_g$. Let $p_{k^*}$ denote the current frontier. Let $T$ denote the total time-steps in the current episode. 
For simplicity, we assume the stochastic goal-reaching policy to be the policy of a random walker, and the deterministic goal-reaching policy to be near-optimal in the sense that given a target goal in $p_{k^*}$, and an initial state in the already explored region, it can precisely reach the  target location at the frontier with a high probability($\gg0$). For analysis, we define SkewFit~\cite{skew}, GoExplore~\cite{goexplore}, and our method in the above setting to be the following simplified schemes, and illustrate the same in Fig.~\ref{fig:simpleanalysis}. To clarify, this is a simplified setup for analysis, and is not our experimental setup in section~\ref{sec:exp}. \vspace*{-0.3cm}\footnote{ Due to space constraint in the paper, we link the proofs of all the lemmas in an Appendix file in the project website \url{https://sites.google.com/view/leaf-exploration/home} }
\begin{enumerate}


    \item \textbf{SkewFit-variant:} The model that executes the stochastic policy for the entire length $T$ of the episode to reach goal $z_g$, and sets diverse goals at the start of every episode, as per the SkewFit objective~\cite{skew}.
    \item \textbf{GoExplore-variant:} The model that first chooses a random previously visited intermediate latent goal state from the archive (not necessarily from the frontier $p_{k^*}$), executes a deterministic goal directed policy for say $k$ steps to reach the intermediate goal, and then executes the stochastic policy exploring for the next $T-k$ steps. For analysis, $0<k<k^*$. This is similar in principle to the idea of Go-Explore~\cite{goexplore}\footnote{Here, we refer to the intuition of Phase 1 of Go-Explore, where the idea is to ``select a state from the archive, go to that state, start exploring from that state, and update the archive." For illustration and analysis, we do not consider the exact approach of Go-Explore. In order to ``go to that state" Go-Explore directly resets the environment to that state, which is not allowed in our problem setting. Hence, in Fig.~\ref{fig:simpleanalysis}, we perform the analysis with a goal-directed policy to reach the state. Also, since Go-Explore does not keep track of reachability, we cannot know whether the intermediate goal chosen can be reached in $k$ timesteps. But for analysis, we assume the intermediate goal is chosen from some $p_k$ distribution as described in Section~\ref{sec:reachidea}}. It also sets diverse goals at the start of every episode, as per the SkewFit objective~\cite{skew}.
    \item \textbf{Our method:} The model that first chooses a state from the frontier $p_{k^*}$, executes a deterministic goal directed policy for say $k^*$ steps to reach the frontier, and then executes the stochastic policy exploring for the next $T-k^*$ steps.  It also sets diverse goals at the start of every episode, as per the SkewFit objective~\cite{skew}.
\end{enumerate}

\begin{lemma}
For a random walk with unit step size in 2D (in a plane), the average distance away from a point (say, the origin) traversed in $t$ time-steps is $\sqrt{t}$.
\label{lemma:main}
\end{lemma}

\noindent The subsequent results are all based on the above stated lemma and assumptions. All the proofs are present in the Appendix.
\begin{lemma}
Starting at $z_0$ and assuming the stochastic goal-reaching policy to be a random walk, on average the SkewFit scheme will reach as far as $R_1 =\sqrt{T}$ at the most in one episode.
\end{lemma}
\begin{lemma}
Starting at $z_0$ and assuming the deterministic goal-deterministic policy to be near optimal (succeeds with a very high probability$>>0$), and the stochastic goal-reaching policy to be a random walk, on average GoExplore scheme will be bounded by $R_2 =\frac{1}{k^*}(\sum_{k=0}^{k^*}(k + \sqrt{T-k})$ in one episode.
\label{lemma:goexplore}
\end{lemma}


\begin{lemma}
Starting at $z_0$ and assuming the deterministic  goal-conditioned policy to be near optimal (succeeds with a very high probability $>>0$), and the stochastic goal-reaching policy to be a random walk, on average our method will be bounded by $R_3 = k^* + \sqrt{T-k^*}$
\end{lemma}
\begin{theorem}
$R_3>R_2$ $\forall k^*\in[1,T)$ and $R_3>R_1$ $\forall k^*\in[1,T)$. i.e. After the frontier $k^*$ is sufficiently large, the maximum distance from the origin reached by our method is more than both GoExplore and SkewFit.
\label{lemma:final}
\end{theorem}

\noindent Theorem~\ref{lemma:final} is the main result of our analysis and it intuitively suggests the effectiveness of deep exploration guaranteed by our method. This is shown through a visualization in Fig.~\ref{fig:simpleanalysis}.

\section{Experiments}
\label{sec:exp}
Our experiments aim to understand the following questions: (1) How does LEAF compare to existing state-of-the-art exploration baselines, in terms of faster convergence and higher cumulative success? 
(2) How does LEAF scale as task complexity increases?
(3) Can LEAF scale to tasks on a real robotic arm?\footnote{Qualitative results of robot videos, and details about the environments and training setup are in an appendix file in the project website (due to space constraint) \href{https://sites.google.com/view/leaf-exploration}{https://sites.google.com/view/leaf-exploration}}

\subsection{Environments}
We consider the image-based and state-based environments illustrated in the supplementary video. In case of image-based environments, the initial state and the goal are both specified as images. In case of state-based environments, the initial state and the goal consists of the coordinates of the object(s), the position and velocity of different components of the robotic arm, and other details as appropriate for the environment (Refer to the Appendix for details). In both these cases, during training, the agent does not have access to any oracle goal-sampling distribution. 
All image-based environments use \textit{only} a latent distance based reward during training (reward $r=-||z_t-z_g||_2$; $z_t$ is the agent's current latent state and $z_g$ is the goal latent state), and have no access to ground-truth simulator states.

\begin{figure*}[h!]
\centering
\hspace*{-0.2cm}
 \begin{subfigure}[b]{0.64\textwidth}
               \centering
    \includegraphics[width=\textwidth]{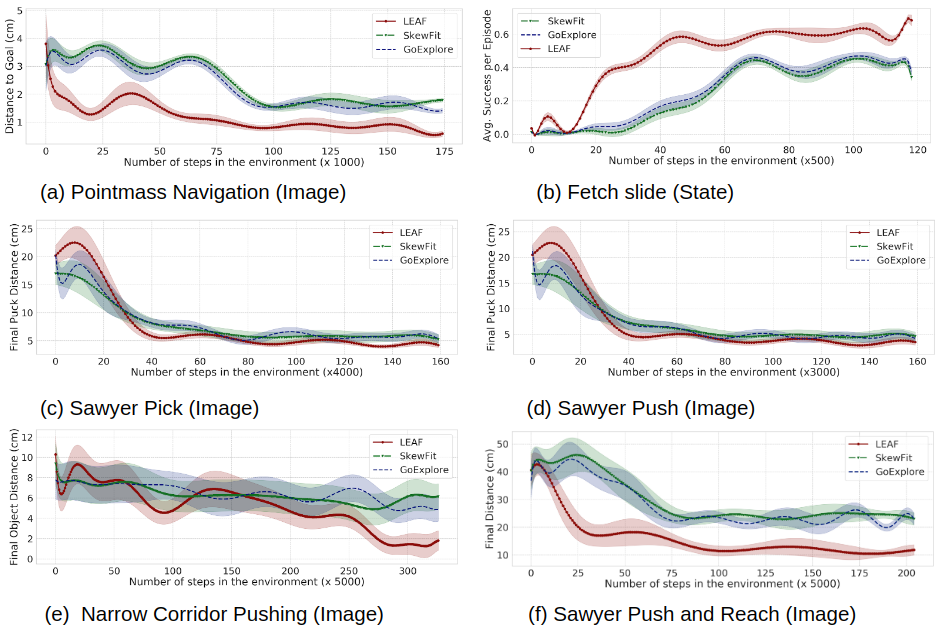}
    \caption{Comparative results on simulation environments}
    \label{fig:quantitative}
        \end{subfigure}\hspace*{-0.05cm}
                 \begin{subfigure}[b]{0.37\textwidth}
               \centering
    \includegraphics[width=\textwidth]{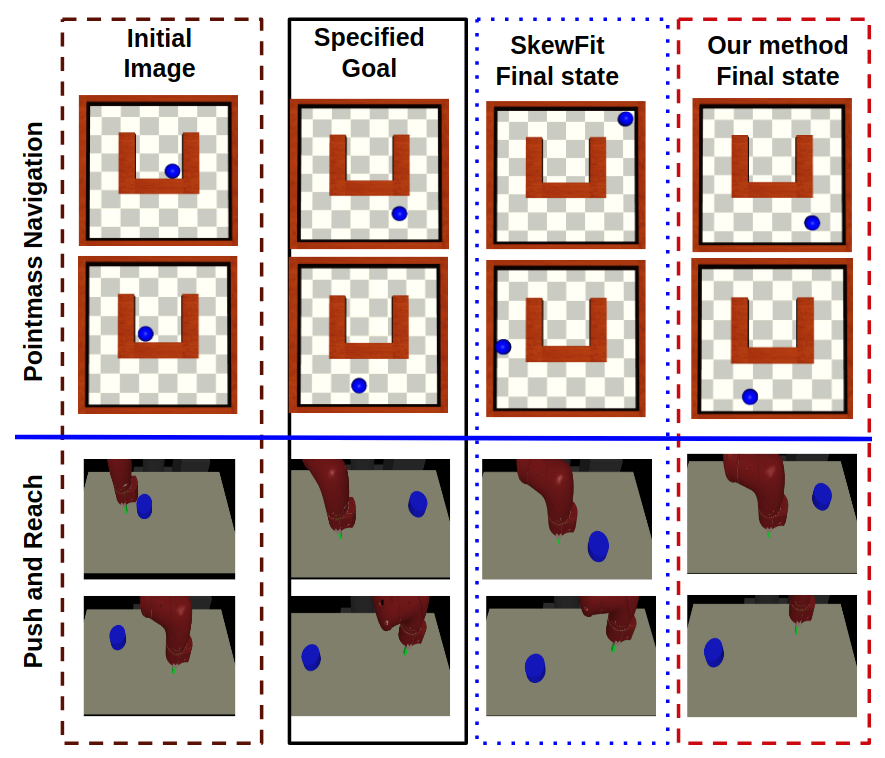}
    \caption{Qualitative results showing the goal image on the left, and rendered images of the final states reached by SkewFit and our method for two evaluation runs in two different environments.}
    \label{fig:qualitative}
        \end{subfigure}

     \caption{Comparison of LEAF against the baselines. The error bars are with respect to three random seeds. The results reported are for evaluations on test suites (initial and goal images are sampled from a test distribution unknown to the agent) as training progresses. Although the ground-truth simulator states are not accessed during training (for (a), (c), (d), (e), and (f)), for the evaluation reported in these figures, we measure the ground truth $L_2$ norm of the final distance of the object/end-effector (as appropriate for the environment) from the goal location.  In (f), we evaluate the final distance between the (puck+end-effector)'s location and the goal location. In (a), (c), (d), (e), and (f), lower is better. In (b), higher is better.}  
 \label{fig:sim_results}
 \vspace*{-0.4cm}
\end{figure*}


\textbf{Image-based environments.} In the \textit{Pointmass Navigation} environment, a pointmass object must be navigated from an initial location to a goal location in the presence of obstacles. The agent must plan a temporally extended path that would occassionally involve moving away from the goal. In \textit{Sawyer Pick}, a Sawyer robot arm must be controlled to pick a puck from a certain location and lift it to a target location above the table. In \textit{Sawyer Push}, a Sawyer robot must push a puck to a certain location with the help of its end-effector. In \textit{Sawyer Push and Reach}, a Sawyer robot arm is controlled to push a puck to a target location, and move its end-effector to a (different) target location. In \textit{Narrow Corridor Pushing}, a Sawyer robot arm must push a puck from one end of a narrow slab to the other without it falling off the platform. This environment is challenging because it requires \textit\textit{{committed exploration}} to keep pushing the puck unidirectionally across the slab without it falling off.

\textbf{State-based environment.} In \textit{Fetch Slide}, a Fetch robot arm must be controlled to push and slide a puck on the table such that it reaches a certain goal location. It is ensured that sliding must happen (and not just pushing) because the goal location is beyond the reach of the arm's end-effector. In \textit{Franka Panda Peg in a Hole}, a real Franka Panda arm must be controlled to place a peg in a hole. The agent is velocity controlled and has access to state information - the position $(x,y,z)$ and velocity $(v_x,v_y,v_z)$ of its end-effector.

\vspace*{-0.13cm}
\subsection{Setup}
We use the Pytorch library~\cite{pytorch1,pytorch2} in Python for implementing our algorithm, and ADAM~\cite{adam} for optimization.  We consider two recent state-of-the-art exploration algorithms, namely Skew-Fit~\cite{skew}, and Go-Explore~\cite{goexplore} as our primary baselines for comparison. 

\vspace*{-0.13cm}
\subsection{Results}
\vspace*{-0.13cm}
\noindent\textbf{Our method achieves higher success during evaluation in reaching closer to the goals compared to the baselines}
Fig.~\ref{fig:sim_results} shows comparisons of our method against the baseline algorithms on all the robotic simulation environments. We observe that our method performs significantly better than the baselines on the more challenging environments. The image based Sawyer Push and Reach environment is challenging because there are two components to the overall task. The Narrow corridor pushing task requires \textit{committed exploration} so that the puck does not fall off the table. The pointmass navigation environment is also challenging because during evaluation, we choose the initial position to be inside the U-shaped arena, and the goal location to be outside it. So, during training the agent must learn to set difficult goals outside the arena and must try to solve them.


In addition to the image based goal environments, we consider a challenging state-based goal environment as proof-of-concept to demonstrate the effectiveness of our method. In Fetch-Slide Fig.~\ref{fig:quantitative} in order to succeed, the agent must acquire an implicit estimate of the friction coefficient between the puck and the table top, and hence must learn to set goals close and solve them before setting goals farther away.

\noindent\textbf{Our method scales to tasks of increasing complexity}
From Fig.~\ref{fig:quantitative} (d. Sawyer Push), we see that our method performs comparably to its baselines. The improvement of our method over SkewFit is not statistically significant. However, in the task of Push and Reach (Fig.~\ref{fig:quantitative}) we observe significant improvement of our method over SkewFit and GoExplore. The baselines succeed either in the push part of the task or the reach part of the task but not both, suggesting overly greedy policies without long term consideration (Refer qualitative results in Fig.~\ref{fig:qualitative}).This suggests the effectiveness of our method in scaling to a difficult task that requires more temporal abstraction. This benefit is probably because our method performs two staged exploration, which enables it to quickly discover regions that are farther away from the initial state given at the start of the episode. 

We observe similar behaviors of scaling to more complicated tasks by looking at the results of the Pointmass Navigation in Fig.~\ref{fig:quantitative}. While the task can be solved even by simple strategies when the goal location is within the U-shaped ring, the agent must actually discover temporally abstracted trajectories when the goal location is outside the U-shaped ring. Discovering such successful trajectories in the face of only latent rewards, and no ground-truth reward signals further demonstrates the need for committed exploration.

\begin{figure}
   \centering
    \includegraphics[width=0.99\columnwidth]{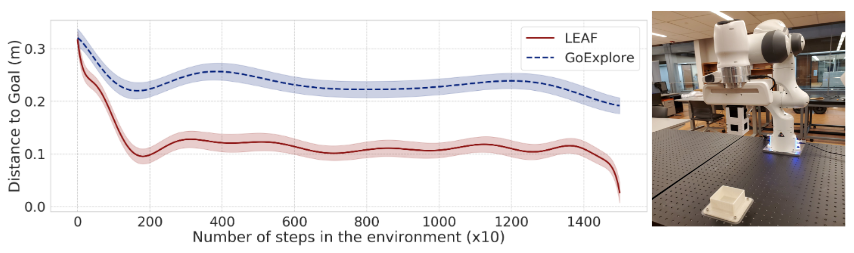}    
    \caption{Results of a Peg-in-a hole task on the real Franka Emika Panda arm. The evaluation metric on the y-axis is the distance ($L_2$ norm) of the center of the bottom of the hole from the center of the bottom of the peg.}
    \label{fig:panda}
     \vspace*{-0.8cm}
  \end{figure}

\noindent\textbf{Our method scales to tasks on a Real Franka Panda robotic arm} Fig.~\ref{fig:panda} shows results for our method on a peg-in-a-hole evaluated on the Franka Panda arm. Here the objective is to insert a peg into a hole, an illustration of which is shown in the supplementary video.  The reward function is defined to be the negative of the distance from the center of bottom of the peg to the center of the bottom of the hole. There is an additional heavy reward penalty for `collision' and a mild penalty for `contact.' The home position of the panda arm (to which it is initialized) is such that the distance from the center of the bottom of the peg to the center of the bottom of the hole is 0.35 meters. The heavy penalty for collision is -10 and the mild penalty for contact is -1. We can observe from Fig.~\ref{fig:panda} that our method successfully solves the task while the baseline GoExplore does not converge to the solution. This demonstrates the applicability of our method on a real robotic arm.


 \vspace*{-0.42cm}
\section{Conclusion} 
 \vspace*{-0.13cm}
\label{sec:conclusion}
In this paper we proposed an algorithm for committed exploration, by keeping track of the frontier of reachable states, and during the start of every episode, executing a deterministic goal-conditioned policy to reach the current frontier, followed by executing a stochastic goal-conditioned exploration policy to reach the goal. The proposed approach can work directly over image-based observations and goal specifications, does not require any reward signal from the environment during training, and is completely self-supervised in that it does not assume access to the oracle goal-sampling distribution for evaluation. Through experiments on seven environments with varying characteristics, task complexities, and temporal abstractions, we demonstrate the efficacy of the proposed approach over state of the art exploration baselines.



\bibliographystyle{ieee}
\bibliography{neurips/bib}
\clearpage
\newpage

\newpage
\section*{Appendix}
\label{sec:appendix}

\section{Proofs of Lemmas}
\begin{figure*}
    \centering
    \includegraphics[width=1.3\columnwidth]{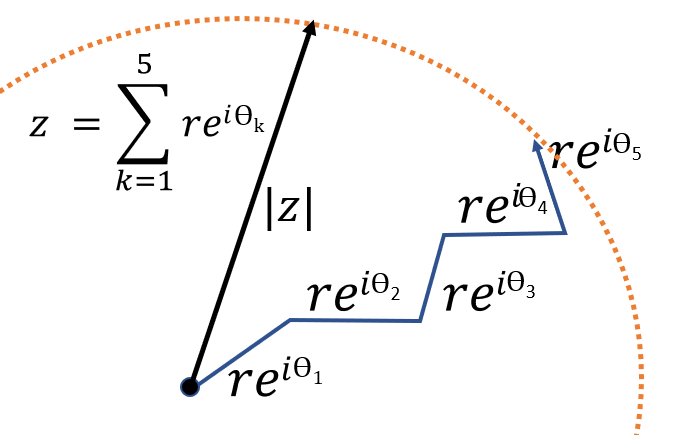}
    \caption{The setting for the proof of Lemma~\ref{lemma:main}. The blue jagged line denotes a random walk in 2D free space where the agent takes a step of size $r$ (We take $r=1$ for convenience and without any loss of generality of the result) at every timestep in any direction $\theta$. So, $re^{i\theta_k}$ is the random variable in the complex plane corresponding to the $k^{th}$ step. }
    \label{fig:my_label}
\end{figure*}
\subsection{Lemma~\ref{lemma:main}}
\begin{proof}
In a plane, consider the sum of $t$ 2D vectors with random orientations. For convenience, we use the phasor notation, and so assume the phase of each vector to be random. Starting at some point (let's call it the origin $z_0$), we assume $t$ steps are taken in an arbitrary direction, which is equivalent to assuming that the angle $\theta$ of the phasors is uniformly distributed in $[0,2\pi)$. The position $z_t$ in the complex plane after $t$ steps is:
$$ z_t = \sum_{k=1}^te^{i\theta_k}$$
The square of the absolute value of $z_t$is:
$$ |z_t|^2 = \sum_{k=1}^te^{i\theta_k}\sum_{m=1}^te^{-i\theta_m}
= t + \underset{k=1,m=1,k\neq m}{\sum\sum^t}e^{i(\theta_k-\theta_m)}$$
Now, we calculate the expected value of the $|z_t|^2$ variable by noting that since $\theta_k$ and $\theta_m$ are randomly sampled in $[0,2\pi)$ with uniform probability, the displacements $e^{i\theta_k}$ and $e^{i\theta_m}$ are random variables, so their difference corresponding to $\theta_k-\theta_m$ is also a random variable and has an expected value (mean) of $0$. Hence,
$$\mathbb{E}_{\theta}|z_t|^2 = \mathbb{E}_{\theta_k,\theta_m\sim[0,2\pi)}\left(t+ \underset{k=1,m=1,k\neq m}{\sum\sum^t}e^{i(\theta_k-\theta_m)} \right)$$
$$\mathbb{E}_{\theta}|z_t|^> = t \implies |z_t|_{rms} = \sqrt{t}$$

So, the average (root-mean-square) distance traversed after $t$ steps starting at some $z_0$ in the 2D plane is $\sqrt{t}$
\end{proof}

\subsection{Lemma 5.2}

\begin{proof}
This follows directly from Lemma~\ref{lemma:main} by setting $t=T$, where $T$ is the max. number of timesteps in the episode.
\end{proof}
\subsection{Lemma 5.3}

\begin{figure*}[t!]
\centering
   \hspace*{-1cm}
       \adjustbox{valign=t}{ \begin{subfigure}[t]{0.80\textwidth}
               \centering
    \includegraphics[width=\textwidth]{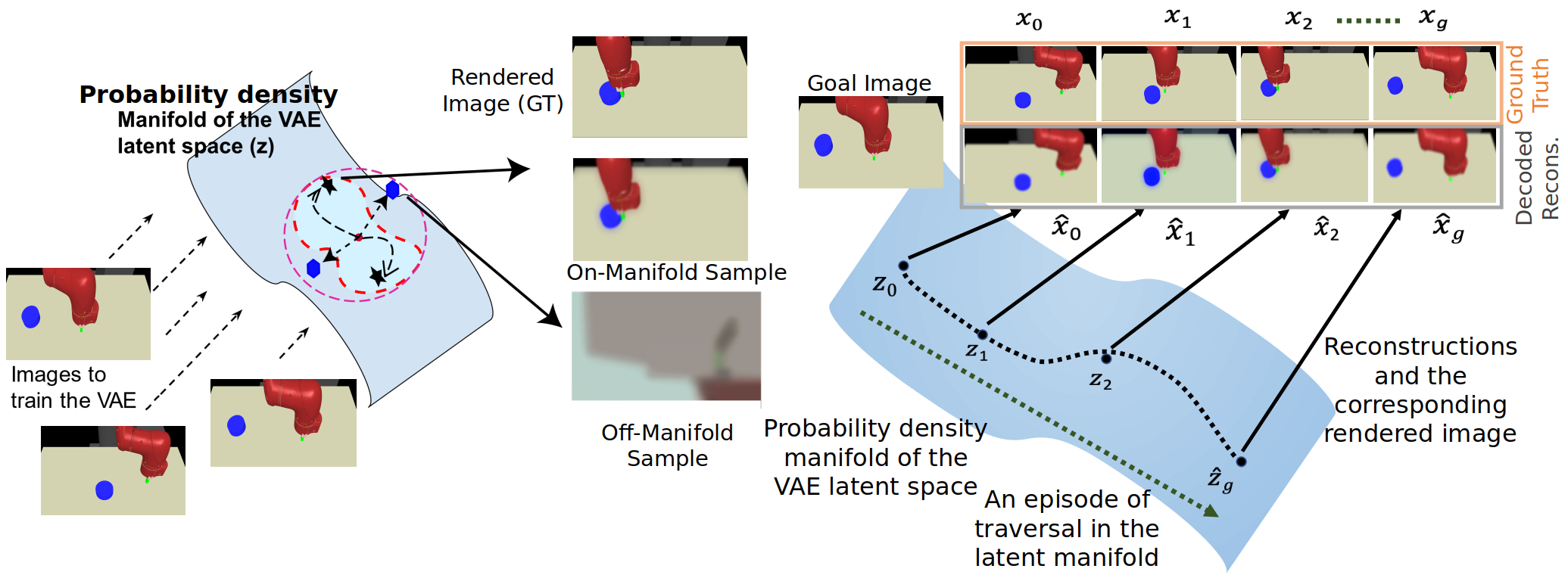}
    \caption{\begin{footnotesize}(1) Our method samples latent goals from the the probability density manifold of the learned VAE latent space $p_\phi(z)$. Sampling states from outside the manifold (i.e. sampling $z$ with low probability density under $p_\phi(z)$) is likely to correspond to images (upon decoding) that do not correspond to a dynamically feasible image of a simulator state. For example, in this figure, given some latent state in the manifold (shown by the center of the sphere), states outside the manifold correspond to regions of the sphere (outlined in pink) that do not intersect the manifold (the intersection is outlined in red). (2) The second figure shows how sampling goals that lie on the latent VAE manifold and traversing on the latent manifold leads to latent states that decode to realistic dynamically feasible image frames.\end{footnotesize}}
    \label{fig:vaegeneral}
        \end{subfigure}}

        \hspace{0.2cm}
        \adjustbox{valign=t}{\begin{subfigure}[t]{0.78\textwidth}
               \centering
   \includegraphics[width=\textwidth]{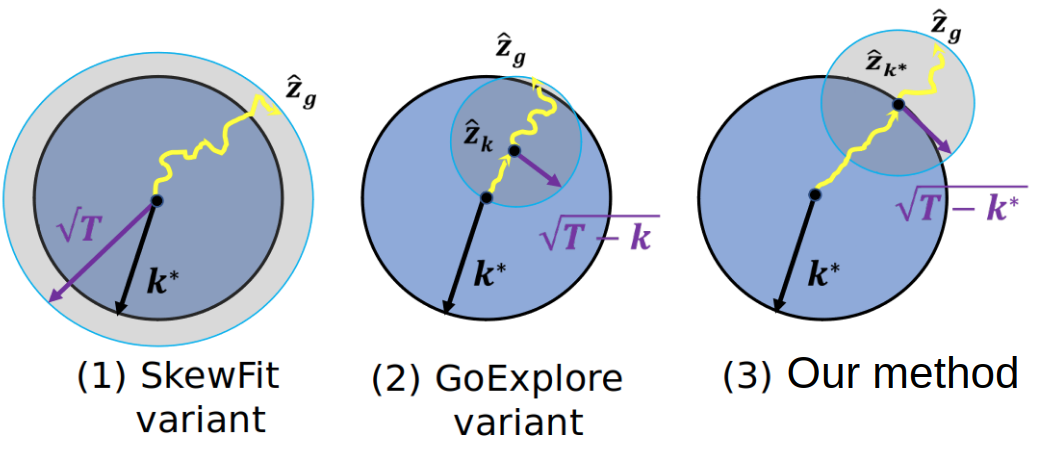}
    \caption{\begin{footnotesize}\textcolor{black}{The blue area of radius $k^*$ denotes the already sufficiently explored region such that $k^*$ is the current frontier. $T$ is the max. number of timesteps of the episode. (3) shows the case when  the deterministic goal reaching policy takes the agent to a state at the frontier in $k^*$ steps, and after that it executes the stochastic policy for $T-k^*$ steps. (2) shows the case when the agent first goes to a random previously visited state, by executing a deterministic goal reaching policy (say, for $k$ steps), and then executes the stochastic policy for $T-k$ steps. (1) shows the case when the agent executes the stochastic policy for $T$ steps starting at the origin. The simplified Go-Explore~\cite{goexplore} and SkewFit~\cite{skew} models are described in Section~\ref{sec:analysis}}\end{footnotesize}}
    \label{fig:analysis1}
          \end{subfigure}}   
         \hspace*{-0.9cm}
    \caption{(a) Sampling from the VAE probability density manifold and traversal in the VAE latent space. (b) Simplified analysis of Our method.}
    \label{fig:vae}
    \vspace*{-0.5cm}
\end{figure*}

\begin{proof}
Since GoExplore first goes to a random previously visited state using the deterministic goal conditioned policy, from Fig~\ref{fig:analysis1} it is evident that the distance of such a state from the start state could be in the range $(0,k^*)$. Let this distance be $k$. Then, from Lemma~\ref{lemma:main}, the max distance on average traversed by the agent from $z_0$, if the intermediate state is at a distance $k$, can be given as:
$$ r_k = k + \sqrt{T-k}$$
Averaging over all the distance of all possible intermediate positions $k\in(0,k^*)$, we obtain the final averaged maximum distance reached by the agent in one episode as:
$$R_2 = \frac{1}{k^*}\sum_{k=0}^{k^*}(k+\sqrt{T-k})$$
\end{proof}
\subsection{Lemma 5.4}
\begin{proof}
In the proof for lemma~\ref{lemma:goexplore}, setting $k=k^*$ in the formula, for $r_k$ gives us the maximum average distance for our method as:
$$R_3 =  k^* + \sqrt{T-k^*}$$
\end{proof}

\subsection{Theorem~\ref{lemma:final}}
\begin{proof}
Given $R_3=k^*+\sqrt{T-k^*}$ and $R_1=\sqrt{T}$. Let us first consider the difference $R_3^2-R_1^2$ below:
\begin{align*}
    R_3^2-R_1^2 &= (k^*+\sqrt{T-k^*})^2 - (\sqrt{T})^2\\
    &= (k^*)^2 + T-k^* = 2k^*\sqrt{T-k^*} - T\\
    &= k^*(k^*-1 + 2k^*\sqrt{T-k^*})\\
    &>0 \quad \forall k*\in(1,T)
\end{align*}
Hence,
$$ R_3^2-R_1^2>0 \quad \forall k*\in(1,T)$$
$$\implies R_3 > R_1  \quad \forall k*\in(1,T)$$

Now, let us consider $R_2 = \frac{1}{k^*}\sum_{k=0}^{k^*}(k+\sqrt{T-k})$. For computing this, we can consider the change of variable $k \implies r$ such that $T-k=r^2$. So, $k=T-r^2$. Hence we have the following relation:
$$R_2 = \frac{1}{k^*}\sum_{r=\sqrt{T-k^*}}^{\sqrt{T}}(T-r^2+r) $$

We have,
$$\sum_{r=\sqrt{T-k^*}}^{\sqrt{T}}(T-r^2+r) =T(\sqrt{T}-\sqrt{T-k^*}) - \sum_{r=\sqrt{T-k^*}}^{\sqrt{T}}(r^2-r)$$

Now, 
\begin{align*}
    \sum_{r=\sqrt{T-k^*}}^{\sqrt{T}}r^2 &=   \sum_{r=0}^{\sqrt{T}}r^2 -   \sum_{r=0}^{\sqrt{T-k^*}}r^2\\
    &= \frac{1}{6}[\sqrt{T}(\sqrt{T}+1)(2\sqrt{T}+1) \\&- \sqrt{T-k^*}(\sqrt{T-k^*}+1)(2\sqrt{T-k^*}+1)] \\
    &= \frac{1}{6}[2T\sqrt{T} +3T + \sqrt{T} \\&- 2(T-k^*)\sqrt{T-k^*} - 3(T-k^*) - \sqrt{T-k^*}]\\
     &= \frac{1}{6}[2T\sqrt{T}  + \sqrt{T} \\&- 2(T-k^*)\sqrt{T-k^*} +3k^* - \sqrt{T-k^*}]
\end{align*}

\begin{align*}
      \sum_{r=\sqrt{T-k^*}}^{\sqrt{T}}r &= \frac{1}{2}[\sqrt{T}(\sqrt{T}+1)-\sqrt{T-k^*}(\sqrt{T-k^*}+1) ]\\
      &= \frac{1}{2}[T + \sqrt{T}- (T-k) - \sqrt{T-k^*}]\\
       &= \frac{1}{2}[ \sqrt{T} +k^* - \sqrt{T-k^*}]
\end{align*}
So, finally,
\begin{align*}
    6\sum_{r=\sqrt{T-k^*}}^{\sqrt{T}}(r^2-r) &= 2T\sqrt{T}  + \sqrt{T} - 2(T-k^*)\sqrt{T-k^*} \\  &+3k^* - \sqrt{T-k^*} - \sqrt{T} -k^* + \sqrt{T-k^*}\\
    &=2T\sqrt{T} -2(T-k^*)\sqrt{T-k^*} + 2k^* \\
\end{align*}
Hence, $R_2$ is given by
\begin{align*}
    R_2 &= \frac{1}{6k^*}[4T\sqrt{T}-4T\sqrt{T-k^*}-k^*\sqrt{T-k^*} -2k^* ]
\end{align*}
Considering $R_3-R_2$, we have,
\begin{align*}
   6k^*( R_3-R_2) &= 6(k^*)^2 + (7k^*+4T)\sqrt{T-k^*} - 4T\sqrt{T} +2k^*\\
\end{align*}
Let, $f(k^*)=6k^*( R_3-R_2) = 6(k^*)^2 + (7k^*+4T)\sqrt{T-k^*} - 4T\sqrt{T} +2k^*$. By treating $k^*$ as continuous, we find that the gradient $\frac{\partial f}{\partial k^*} > 0$ $\forall k^*\in(0,T]$. Also, the value $f(k^*=0)=0$. Hence, we have proved that $R_3>R_2$ $\forall k\in[1,T]$.
\end{proof}
\clearpage
\newpage

\clearpage
\newpage

\begin{algorithm}[H]
	\begin{algorithmic}[1]
		\Procedure{Overall}{GCP $\mu_\theta$, $\pi_\theta$, Prior $p(z)$, $S$, $f_\psi(\cdot)$, $IsReachable(\cdot)$, $H_0$, $N$, $\epsilon$, $\delta$, ReplayBuffer $\mathcal{B}$}
		\State $p_\phi = p(z)$
		\State Init $ReachNet(\cdot,\cdot; k)$, Def ReachData $\mathcal{D}$, $K=H_0N$
		\For{episodes = 1 to $N$}
		\State $s_0\sim S_0$, encode $z_0=f_\psi(s_0)$, $z_{g}\sim p_\phi,\hat{z}^*=z_0$
	\If{episodes $\geq$ $N/4$} 
	 \For{$k=K-1,...,1$}
	\State Set the value of $IsReachable(z_0;k)$
	 \EndFor
	
	  \State  $k^*=\argmax_k(IsReachable(z_0;k))$ 
	  \State Choose $ z_{k^*} = \argmin_{z\sim p_{k^*}}||z-z_g||_2$
	  	\State Exec $\pi_\theta(z_0;z_{k^*})$, reach $\hat{z}^*$ s.t. $||\hat{z}^*-z_{k^*}||\approx 0$
	  	\EndIf
		\State Set $H=H_0*episodes$ // \texttt{Curriculum} 
		\For{$t=0,...,H-1$}
		\If{$t==0$}
		\State Select $a_t\sim\mu_\theta(\hat{z}^*;z_{g})$ and execute it
		\EndIf
		\If{$t>0$ and $t<k^*$}
		\State Select $a_t\sim\mu_\theta(f_\psi(s_t);z_{g})$ and execute it
		\EndIf 
			\If{$t>0$ and $t\geq k^*$}
		\State Select $a_t\sim\pi_\theta(f_\psi(s_t);z_{g})$ and execute it
		\EndIf 
		\State // \texttt{Store tuple and Train policy}
		\State \textsc{StoreandTrain()}
		\EndFor 
\State \textsc{AppendtoDataset($\mathcal{D}$)}

\State \textsc{SampleFutureStates($\mathcal{B}$)}
    \State Construct skewed dist.  $p_{skewed}$ from $\mathcal{B}$ 
    \State Train the encoder using $p_\phi = p_{skewed}$
    \State Train $ReachNet(\cdot,\cdot; k)$ using ReachDataset $\mathcal{D}$
		\EndFor
		\EndProcedure
	\end{algorithmic}
	\caption{Overview of the proposed Algorithm}
	\label{alg:mainbrief}
\end{algorithm}

\begin{algorithm}[h!]

	\begin{algorithmic}[1]
		\Procedure{StoreandTrain}{()}
			\State The env transitions to state $s_{t+1}\sim p(\cdot|s_t,a_t)$
		\State Store $(s_{t},a_t,s_{t+1},z_g,t+1)$ in replay buffer $\mathcal{R}$
		\State Sample transition $(s,a,s',z^h_g,k)\sim \mathcal{R}$
		\State Encode $z = f_\psi(s), z' = f_\psi(s')$
		\State Compute the latent reward $r=-||z'-z^h_g||$
		\State Minimize Bellman Error with $(z,a,z',z^h_g,k,r)$
    		\EndProcedure
		\Procedure{AppendtoDataset}{ReachDataset $\mathcal{D}$}
		\For{$t_1 = 0,...,H-1$}
\For{$t_2 = 0,...,H-1$}
\If{$t_1\neq t_2$}
\State ReachDataset.add($(z_{t_1},z_{t_2},|t_1-t_2|)$)
\EndIf
\EndFor
\EndFor
		\EndProcedure
		\Procedure{SampleFutureStates}{$\mathcal{B}$}
			\For{$t=0,...,H-1$}
	\For{$i=0,...,k-1$}
    \State Sample future states $s_{h_i}, t\leq h_i-1 \leq H-2$
    \State Store $(s_t,a_t,s_{t+1},f_\psi(s_{h_i}),h_i+1)$ in $\mathcal{R}$
    	\EndFor
    		\EndFor
    		\EndProcedure
	\end{algorithmic}
	\caption{Subroutines of the algorithm (forward referenced in Alg 1.)}
	\label{alg:main}
\end{algorithm}
\section{Details about the environments}

 \begin{figure}[h!]
     \centering
  
     \includegraphics[width=0.99\columnwidth]{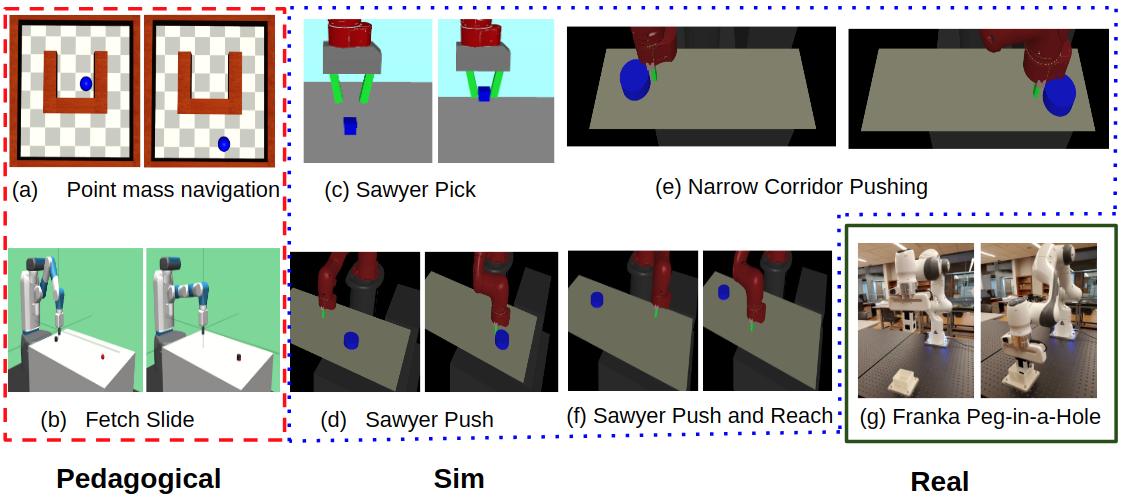}
   
     \caption{Illustrations of all the environments evaluated in this paper, with example initial (left) and goal (right) images.  Each environment we consider is meant to demonstrate scaling of the proposed approach to different settings: image-based goals, state-based goals, continuous action spaces, different dynamics (point mass, Fetch robot, Sawyer robot, Franka Panda robot), increasing task complexity (Push to Push and Reach, and to Pushing across a narrow corridor), and applicability to a real robot. Specific details about each environment are in the Appendix. }
     \label{fig:sim_envs}
     
     \vspace*{-0.4cm}
 \end{figure} 
 
\begin{itemize}
  \item \textbf{Pointmass Navigation} (\textit{Image-based}): This is a 2D environment where an object (circular with diameter 1cm) must learn to navigate a square-shaped arena from an initial location to a goal location in the presence of obstacles. The obstacle is a U-shaped wall in the  center. The agent must plan a temporally extended path that would occassionally involve moving away from the goal. This is a 2D environment with a continuous action space, and the agent has access to images from a top-down view. The observations and goals for evaluation are specified as 48x48 RGB images. The total dimensions of the arena are 8cmx8cm, and the thickness of the U-shaped wall is 1cm. The action space is a 2D vector $(v_x,v_y)$ consisting of velocity components in the x and y dimensions. For evaluation, the final $(x,y)$ position of the center of the object at the end of the episode is computed from the actual $(x,y)$ position of the goal. During training, the only reward is the distance (L2 norm) between the current latent state $z'$ and the latent goal state$z^h_g$, $r=-||z'-z^h_g||$ as shown in the Algorithm.
   \item \textbf{Fetch Slide} (\textit{State-based}): A Fetch robot arm must be controlled to push and slide a puck on the table such that it reaches a certain goal location. It is ensured that sliding must happen (and not just pushing) because the goal location is beyond the reach of the arm's end-effector. We make this environment significantly difficult by reducing the default friction coefficient by a factor of 2, to have $\mu=0.3$. The workspace has a dimension of 10cmx30cm. Since, we assume access to the state of the simulator here, goals are specified as the desired $(x,y)$ position of the puck. The observations are specified as a vector containing the absolute position of the gripper, and the relative position of the object and target goal position. Actions are 3D vectors specifying the desired relative gripper position at the next timestep. Rewards during training are sparse and they do access the ground-truth simulator states (unlike the image based environments where we use latent rewards): $r(s,a,g)=-\mathbb{I}\{|g-s'|\leq \epsilon\}$, where $\mathbb{I}$ is the indicator function, $g$ denotes the goal state, $s'$ denotes the next state after executing action $a$ in state $s$.
     \item \textbf{Sawyer Pick} (\textit{Image-based}): A MuJoCo powered 7-DoF Sawyer robot arm must be controlled to pick a puck from a certain location and lift it to a target location above the table. The observations and goals for evaluation are specified as 84x84 RGB images. The workspace dimensions of the table are 10cmx15cm ($x$ and $y$ dimensions). However, the robot is constrained to move only in the $y$ and $z$ dimensions. The dimension that can be accessed in the $yz$ plane is 15cmx15cm. The object to be picked up is a cube of edge length 1.5cm. The robot is position controlled and, action space is continuous and is a vector $(a_x,a_z, a_{grip})$ denoting relative position of the end-effector at the next timestep in $yz$ plane, and the separation between the gripper fingers. For evaluation, the final $(y,z)$ position of the center of the object at the end of the episode is computed from the actual $(y,z)$ position of the object in the goal image. During training, the only reward is the distance (L2 norm) between the current latent state $z'$ and the latent goal state $z^h_g$, $r=-||z'-z^h_g||_2$ as shown in the Algorithm.
         \item \textbf{Sawyer Push} (\textit{Image-based}): A MuJoCo powered 7-DoF Sawyer robot must push a puck to a certain location with the help of its end-effector. The observations and goals for evaluation are specified as 84x84 RGB images. The workspace dimensions of the table are 20cmx30cm. The puck has a radius of 6cm. The robot is position controlled, the action space is continuous and is a vector $(a_x,a_y)$ denoting relative position of the end-effector at the next timestep in $xy$ plane. For evaluation, the final $(x,y)$ position of the center of the object at the end of the episode is computed from the actual $(x,y)$ position of the puck in the goal image. During training, the only reward is the distance (L2 norm) between the current latent state $z'$ and the latent goal state $z^h_g$, $r=-||z'-z^h_g||$ as shown in the Algorithm.
    \item \textbf{Sawyer Push and Reach} (\textit{Image-based}): A MuJoCo powered 7-DoF Sawyer robot arm must be controlled to push a puck to a target location, and move its end-effector to a (different) target location. The observations and goals for evaluation are specified as 84x84 RGB images. The workspace dimensions of the table are 25cmx40cm. The puck has a radius of 6cm. The robot is position controlled, the action space is continuous and is a vector $(a_x,a_y)$ denoting relative position of the end-effector at the next timestep in $xy$ plane. For evaluation, the concatenation of the final $(x,y)$ position of the center of the puck, and the  $(x,y)$ position of the end-effector at the end of the episode is computed from the actual $(x,y)$ position of the center of the puck, and the  $(x,y)$ position of the end-effector in the goal image. During training, the only reward is the distance (L2 norm) between the current latent state $z'$ and the latent goal state $z^h_g$, $r=-||z'-z^h_g||$ as shown in the Algorithm.
  \item \textbf{Narrow corridor pushing} (\textit{Image-based}): A MuJoCo powered 7-DoF Sawyer robot arm must be controlled to push a puck to a target location through a narrow corridor wwithout the puck falling off the table. The observations and goals for evaluation are specified as 84x84 RGB images. The workspace dimensions of the table are 18cmx10cm. The puck has a radius of 3cm. The robot is position controlled, the action space is continuous and is a vector $(a_x,a_y)$ denoting relative position of the end-effector at the next timestep in $xy$ plane. For evaluation, the final $(x,y)$ position of the center of the object at the end of the episode is computed from the actual $(x,y)$ position of the puck in the goal image. During training, the only reward is the distance (L2 norm) between the current latent state $z'$ and the latent goal state $z^h_g$, $r=-||z'-z^h_g||$ as shown in the Algorithm.
    \item \textbf{Franka Panda Peg in a Hole} (\textit{Image-based; Real robot}); A real Franka Panda arm must be controlled to place a peg in a hole.  The agent is velocity controlled and has access to state information of the robot. The states observed at each time-step are 6-tuples containing the positions and velocities of the end-effector $(x,y,z,v_x,v_y,v_z)$. The reward function is defined to be the negative of the distance from the center of bottom of the peg to the center of the bottom of the hole. There is an additional heavy reward penalty for `collision' and a mild penalty for `contact.' The home position of the panda arm (to which it is initialized) is such that the distance from the center of the bottom of the peg to the center of the bottom of the hole is 0.35 meters. The heavy penalty for collision is -10 and the mild penalty for contact is -1. 
\end{itemize}

\section{Implementation Details}
We use a Dell ALIENWARE AURORA R8 desktop for all experiments. It has 2x GTX 1080 Ti 11GB GPUs and 9th Gen Intel Core i7 9700K (8-Core/8-Thread, 12MB Cache, Overclocked up to 4.6GHz on all cores).

\begin{itemize}
    \item \textbf{Hyperparameters}: The hidden sizes of the Q-network and the Policy Network are both [400,300]. The Activation functions in both these networks are ReLU. The Batch size is 128 for the Pointmass Navigation environment and is 20148 for all the other image-based environments. We use ADAM optimizer, learning 0.001 (for all networks), and a Replay Buffer size of 1000000.
    \item \textbf{Reachability Network}: The encoders of the reachability network (both the tied latent state encoders and the encoder for \textbf{k}) are multi-layered perceptrons with sizes [16,102,90,100] and ReLU non-linearities. The joint fully connected layers have sizes [300,80,70,1] with ReLU non-linearities. 
    \item \textbf{VAE}: For the vision-based environments, the kernel sizes, strides, and channels of the encoder respectively are [5,5,5], [3,3,3], [16,16,32]. The  kernel sizes, strides, and channels of the decoder respectively are [5,6,6], [3,3,3], [32,32,16]. 
\end{itemize}


\end{document}